\documentclass[conference]{IEEEtran}
\IEEEoverridecommandlockouts
\usepackage{cite}
\usepackage{amsmath,amssymb,amsfonts}
\usepackage{algorithmic}
\usepackage{graphicx}
\usepackage{textcomp}
\usepackage{xcolor}

\usepackage[ruled]{algorithm2e} 
\usepackage{flushend}
\usepackage{amsmath,amsfonts,amsthm,bm}
\usepackage{stfloats}
\usepackage{graphicx}
\usepackage{textcomp}
\usepackage{xcolor}
\usepackage{booktabs} 

\usepackage{amsmath}
\usepackage{rotating}
\usepackage{booktabs}

\def\BibTeX{{\rm B\kern-.05em{\sc i\kern-.025em b}\kern-.08em
    T\kern-.1667em\lower.7ex\hbox{E}\kern-.125emX}}
\begin{document}

\title{SaliencyDecor: Enhancing Neural Network Interpretability through Feature Decorrelation
}

\author{
Ali Karkehabadi$^{1*}$,
Jamshid Hassanpour$^{2*}$,
Houman Homayoun$^{1}$,
Avesta Sasan$^{1}$\\
$^{1}$University of California, Davis, USA\\
$^{2}$Georgia Institute of Technology, USA\\
\{akarkehabadi, hhomayoun, asasan\}@ucdavis.edu,
jamshid@gatech.edu\\
$^{*}$ These authors contributed equally to this work.
}

\maketitle

\makeatletter
\def\ps@IEEEtitlepagestyle{
  \def\@oddhead{\underline{\hbox to\textwidth{\textit{Preprint. Accepted at IJCNN 2026.}\hfill}}}
  \def\@evenhead{\underline{\hbox to\textwidth{\textit{Preprint. Accepted at IJCNN 2026.}\hfill}}}
  \def\@oddfoot{}
  \def\@evenfoot{}
}
\def\ps@headings{
  \def\@oddhead{\underline{\hbox to\textwidth{\textit{Preprint. Accepted at IJCNN 2026.}\hfill}}}
  \def\@evenhead{\underline{\hbox to\textwidth{\textit{Preprint. Accepted at IJCNN 2026.}\hfill}}}
  \def\@oddfoot{}
  \def\@evenfoot{}
}
\makeatother

\pagestyle{headings}

\begin{abstract}
Gradient-based saliency methods are widely used to interpret deep neural networks, yet they often produce noisy and unstable explanations that poorly align with semantically meaningful input features. We argue that a fundamental cause of this behavior lies in the geometry of learned representations: correlated feature dimensions diffuse attribution gradients across redundant directions, resulting in blurred and unreliable saliency maps. To address this issue, we identify feature correlation as a structural limitation of gradient-based interpretability and propose \textit{SaliencyDecor}, a training framework that enforces feature decorrelation to improve attribution fidelity without modifying saliency methods or model architectures by reshaping the feature space toward orthogonality, our approach promotes more concentrated gradient flow and improves the fidelity of saliency-based explanations. SaliencyDecor jointly optimizes classification, prediction consistency under feature masking, and a decorrelation regularizer, requiring no architectural changes or inference-time overhead. Extensive experiments across multiple benchmarks and architectures demonstrate that our method produces substantially sharper and more object-focused saliency maps while simultaneously improving predictive performance, achieving accuracy gains across the datasets. These results establish our method as a principled mechanism for enhancing both interpretability and accuracy, challenging the conventional trade-off between explanation quality and model performance.
\end{abstract}


\section{Introduction}
\label{sec:intro}

Deep neural networks (DNNs) have achieved state-of-the-art performance across a wide range of tasks, including computer vision, natural language processing, and time series analysis \cite{krizhevsky2012imagenet, devlin2018bert}. Despite their success, the lack of interpretability in these models remains a major obstacle to their deployment in high-stakes applications. Understanding which input features drive model predictions is essential for validating model behavior, diagnosing failures, and establishing user trust \cite{lipton2018mythos, caruana2015intelligible}. Saliency methods address this challenge by assigning importance scores to input features based on their influence on model outputs \cite{baehrens2010explain, sundararajan2017axiomatic}. However, many existing approaches produce explanations that are noisy, unstable, and poorly aligned with human intuition \cite{adebayo2018sanity}. In practice, saliency maps often exhibit weak foreground-background separation and inconsistent attribution of semantically relevant regions, limiting their usefulness for both interpretation and debugging.

A key source of this behavior lies in the structure of the learned feature representations. Gradient-based explanations are particularly sensitive to correlations in the underlying feature space. When multiple features encode redundant information, gradients may diffuse across correlated dimensions, obscuring truly influential input elements and amplifying noise in irrelevant regions. Related phenomena have been studied in representation learning under the notion of dimensional collapse, where learned representations concentrate along a small number of correlated directions despite high nominal dimensionality \cite{ huang2018decorrelated}. Such collapse reduces representation efficiency and degrades gradient signal quality. Standard normalization techniques such as Batch Normalization \cite{ioffe2015batch} correct first-order statistics but do not explicitly remove inter-feature correlations, leaving this issue largely unaddressed.
Several saliency methods attempt to mitigate noisy explanations through post-hoc smoothing or aggregation. SmoothGrad averages gradients over noisy input perturbations \cite{smilkov2017smoothgrad}, while Integrated Gradients accumulates gradients along a path from a baseline input \cite{sundararajan2017axiomatic}. Although effective in some cases, these techniques do not modify the underlying representations and therefore cannot resolve the structural causes of saliency degradation. 

Similarly, saliency-guided training approaches \cite{ismail2021improving} incorporate gradient information into optimization but still operate on correlated feature spaces, resulting in unstable and misleading attributions \cite{kapishnikov2021guided}. Recent works have attempted to improve saliency-guided training through more adaptive masking strategies, such as SMOOT \cite{karkehabadi2024smoot} and HLGM \cite{karkehabadi2024hlgm}, which enhance both accuracy and saliency quality. However, these methods still rely on the underlying feature representations and do not explicitly address feature correlation, leaving structural sources of attribution instability unresolved.

The limitations of current saliency methods have been systematically documented. These findings suggest that improving explanation quality requires addressing structural properties of neural representations rather than modifying gradient computations alone.

In this work, we identify feature correlation in learned representations as a key bottleneck for saliency-based interpretability. Correlated features introduce redundancy, distort importance distributions, and reduce the contrast between relevant and irrelevant regions in saliency maps. We propose \emph{SaliencyDecor}, a training framework that explicitly addresses this issue by integrating ZCA whitening with saliency-guided optimization. By decorrelating intermediate representations during training, our method promotes more balanced gradient propagation and yields saliency maps that are sharper, more stable, and better aligned with semantically meaningful features. Extensive experiments on image classification benchmarks demonstrate consistent improvements in saliency quality, accompanied by competitive or improved classification performance.

\section{Related Work}
\label{sec:related}

\subsection{Saliency Methods}
Gradient-based saliency attributes predictions using input derivatives, beginning with raw gradients \cite{baehrens2010explain} and later refinements\cite{sundararajan2017axiomatic, smilkov2017smoothgrad, shrikumar2017learning, lundberg2017unified }. Despite their popularity, these methods often produce noisy and unstable explanations. Prior studies show that saliency maps can fail sanity checks \cite{adebayo2018sanity} and are highly sensitive to small input perturbations \cite{ghorbani2019interpretation}.

\subsection{Feature Decorrelation}
Decorrelation has been explored to improve representation quality and optimization. DeCov \cite{cogswell2015reducing} penalizes correlated activations, while Decorrelated Batch Normalization \cite{huang2018decorrelated} incorporates whitening into normalization layers. In self-supervised learning, whitening and redundancy-reduction techniques have been shown to mitigate dimensional collapse and improve representation diversity \cite{hua2021feature, zbontar2021barlow}. Recent studies further highlight the prevalence and impact of dimensional collapse in learned representations, demonstrating its effect on feature quality and downstream performance \cite{inproceedings}. However, the connection between feature decorrelation and gradient-based interpretability remains largely unexplored.

\subsection{Saliency-Guided Training and Interpretability}
Several works integrate interpretability objectives into training, including gradient regularization \cite{ross2017right}, explanation supervision \cite{ghaeini2019saliency}, and saliency-guided training \cite{ismail2021improving}. While effective in some settings, these approaches operate on correlated representations and do not address structural sources of gradient noise \cite{kapishnikov2021guided}. Alternative strategies modify architectures or objectives through class activation mapping \cite{zhou2016learning}, distillation \cite{frosst2017distilling}, or robustness-based alignment \cite{ etmann2019connection}.

\subsection{Normalization and Representation Structure}
Normalization methods such as Batch Normalization \cite{ioffe2015batch} stabilize first-order statistics but leave feature correlations largely unchanged, while whitening variants mainly target optimization efficiency \cite{huang2018decorrelated}. In contrast, we explicitly leverage feature decorrelation to improve saliency fidelity, linking representation geometry with attribution quality.

\section{Problem Statement: The Dimensional Collapse Challenge in Interpretability}

Neural network explanations based on input gradients suffer from geometric limitations induced by correlated feature spaces, leading to two forms of collapse. 
{Complete collapse} occurs when representations become constant, yielding $\nabla_X f_\theta(X) \approx 0$ and eliminating informative gradients. 
{Dimensional collapse} arises when representations lie on a low-dimensional manifold, i.e., the covariance $\boldsymbol{\Sigma}$ has low effective rank,
\[
\mathrm{rank}_{\mathrm{eff}}(\boldsymbol{\Sigma}) \ll \mathrm{dim}(\boldsymbol{\Sigma}),
\quad
\mathrm{rank}_{\mathrm{eff}}(\boldsymbol{\Sigma}) = \exp\!\big(-\textstyle\sum_i \tilde{\lambda}_i \log \tilde{\lambda}_i\big),
\]
where $\tilde{\lambda}_i = \lambda_i / \sum_j \lambda_j$ are normalized eigenvalues of $\boldsymbol{\Sigma}$.

In such settings, correlated features cause gradients to diffuse across redundant dimensions rather than concentrate on informative ones, limiting interpretability.
Current methods fail to recognize that the geometric structure of the representation space directly impacts gradient flow and feature attribution. Without proper decorrelation, we show that saliency methods suffer from:

\begin{itemize}
    \item {Gradient Entanglement}: Correlated features produce entangled gradients that distribute attribution across redundant features.
    
    \item {Vanishing Saliency Gaps}: The separation between salient and non-salient features diminishes as correlation increases.
    
    \item {Stochastic Axis Swapping}: During training, correlated representations cause unstable gradient directions.
\end{itemize}

These challenges require a principled approach that transforms the geometry of the feature space to enable faithful gradient-based explanations.

\section{Methodology: SaliencyDecor Framework}

\begin{figure*}[t]
  \centering
  \includegraphics[width=1.5
  \columnwidth]{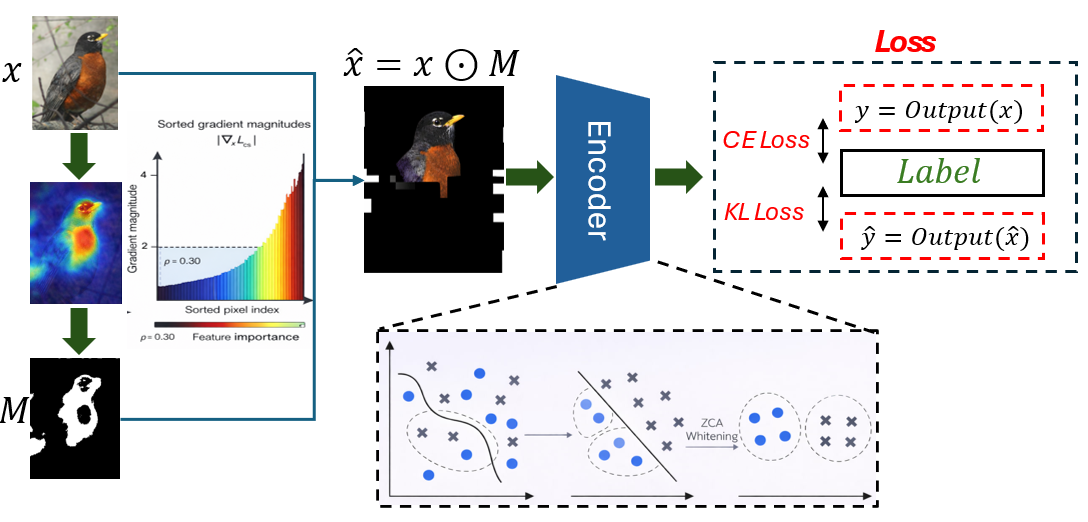}
  \caption{\textbf{Overview of the proposed SaliencyDecor framework.}
Given an input $X$, gradient-based importance scores are used to identify non-informative regions and generate a saliency mask $M$. Intermediate encoder features are decorrelated via group-wise ZCA whitening to reduce redundancy and stabilize gradient attribution. The network is trained with a multi-objective loss combining classification, consistency, and decorrelation terms, encouraging orthogonal feature representations that produce sharper and more faithful saliency maps without inference-time overhead.}
  \label{fig:arch}
\end{figure*}

SaliencyDecor enhances gradient-based interpretability through feature decorrelation during training. Our framework addresses dimensional collapse by combining ZCA whitening with saliency-guided optimization, creating representations where gradients naturally align with human-interpretable features. Figure~\ref{fig:arch} shows the general architecture of our method.

\subsection{Geometric Whitening for Decorrelated Features}

To prevent dimensional collapse, SaliencyDecor applies ZCA whitening. 
Given features $\mathbf{Z} \in \mathbb{R}^{d \times m}$,
\[
\tilde{\mathbf{Z}} = \boldsymbol{\Sigma}^{-1/2}(\mathbf{Z} - \boldsymbol{\mu}\mathbf{1}^T),
\]
where $\boldsymbol{\mu}$ is the batch mean and 
$\boldsymbol{\Sigma} = \tfrac{1}{m}(\mathbf{Z} - \boldsymbol{\mu}\mathbf{1}^T)(\mathbf{Z} - \boldsymbol{\mu}\mathbf{1}^T)^T 
= \mathbf{D}\boldsymbol{\Lambda}\mathbf{D}^T$ is the covariance eigendecomposition, with $\mathbf{D}$ the eigenvectors and $\boldsymbol{\Lambda}$ the eigenvalues, so that $\boldsymbol{\Sigma}^{-1/2} = \mathbf{D}\boldsymbol{\Lambda}^{-1/2}\mathbf{D}^T$.
We choose ZCA because, unlike PCA whitening, it preserves feature orientation while enforcing orthogonality, ensuring gradients flow along consistent semantic directions.

\subsection{Efficient Group-wise Feature Decorrelation}

To address computational challenges in high-dimensional spaces, SaliencyDecor implements efficient group-wise whitening. Features are partitioned into groups of size $G$, and whitening is applied within each group:

\begin{equation}
\mathbf{Z}^{[h]} = \text{ZCA}(\mathbf{Z}^{[h]})
\end{equation}

\noindent where $\mathbf{Z}^{[h]}$ represents the $h$-th group of features. This reduces computational complexity from $\mathcal{O}(d^2m)$ to $\mathcal{O}(dmG)$, making our method scalable to modern architectures with millions of parameters.

\vspace{-0.5pt}
\subsection{Multi-objective Optimization Framework}

Our learning objective combines three components:
\begin{equation}
\mathcal{L} = \mathcal{L}_{\text{cls}} + \alpha \cdot \mathcal{L}_{\text{consistency}} + \lambda \cdot \mathcal{L}_{\text{decorr}}
\end{equation}

\noindent where:
\begin{itemize}
    \item $\mathcal{L}_{\text{cls}}$ is the standard cross-entropy loss for classification
    \item $\mathcal{L}_{\text{consistency}} = \mathcal{D}_{\text{KL}}(f_\theta(X) \| f_\theta(\widetilde{X}))$ is the KL-divergence between predictions from original and masked inputs
    \item $\mathcal{L}_{\text{decorr}} = \|\tilde{\mathbf{Z}}^T\tilde{\mathbf{Z}} - \mathbf{I}\|_F$ measures deviation from orthogonality
    \item $\alpha$ and $\lambda$ control the trade-off between accuracy, consistency, and decorrelation
\end{itemize}


For feature masking, we leverage decorrelated gradients to identify truly non-informative features. Specifically, we compute importance scores from gradients in the whitened space and project them back to the input domain:

\begin{equation}
\text{importance} = |\nabla_X f_\theta(X)|
\end{equation}

This ensures that masking decisions reflect true information content rather than correlation artifacts.

\subsection{SaliencyDecor Algorithm}
In Algorithm \ref{alg:SaliencyDecor}, we present the complete SaliencyDecor procedure. For each minibatch, we first compute feature representations through the encoder. We then apply group-wise ZCA whitening to these features, decorrelating them while preserving their original orientation. After computing the classification loss, we calculate a decorrelation regularization term that encourages orthogonality in feature space.
For creating input masks, we compute gradients of the classification loss with respect to the input and identify the least important features based on gradient magnitude. We then generate a masked input by replacing these features and compute a consistency loss to ensure that the model's predictions remain stable despite feature masking. Finally, we update the parameters using the combined loss that balances classification accuracy, mask consistency, and feature decorrelation. SaliencyDecor is theoretically grounded in geometric deep learning principles, where the structure of the representation space directly impacts gradient flow and feature attribution. By reshaping the geometry of the feature space, SaliencyDecor creates representations where gradient-based saliency naturally aligns with human-interpretable features without requiring post-hoc modifications to existing saliency methods. This foundational approach provides consistent improvements across various architectures and saliency methods, addressing the core geometric limitations that have hindered interpretability in deep learning.

\begin{algorithm}[t]
\caption{SaliencyDecor: Decorrelated Saliency-Guided Training}
\KwIn{Data $\{X,y\}$, mask ratio $\rho$, weights $\alpha,\lambda$, group size $G$}
Initialize parameters $\theta$ \;

\For{each minibatch $\{X,y\}$}{
  $Z \leftarrow f_{\text{enc}}(X)$ \;
  $\tilde{Z} \leftarrow \text{GroupZCA}(Z,G)$ \;

  $\hat{y} \leftarrow f_{\text{cls}}(\tilde{Z})$ \;
  $\mathcal{L}_{\text{cls}} \leftarrow \text{CE}(\hat{y},y)$ \;
  $\mathcal{L}_{\text{decorr}} \leftarrow \|\tilde{Z}^\top \tilde{Z} - I\|_F$ \;

  $S_{\tilde{Z}} \leftarrow |\nabla_{\tilde{Z}} \mathcal{L}_{\text{cls}}|$ \;
  $S_X \leftarrow \text{ProjectToInput}(S_{\tilde{Z}})$ \;
  $\tilde{X} \leftarrow \text{Mask}(X,\rho,S_X)$ \;

  $\mathcal{L}_{\text{cons}} \leftarrow \mathcal{D}_{\mathrm{KL}}\big(f_\theta(X)\,\|\,f_\theta(\tilde{X})\big)$ \;
  $\mathcal{L} \leftarrow \mathcal{L}_{\text{cls}} + \alpha \mathcal{L}_{\text{cons}} + \lambda \mathcal{L}_{\text{decorr}}$ \;

  Update $\theta$ \;
}
\label{alg:SaliencyDecor}
\end{algorithm}





\section{Empirical Validation and Analysis}
We benchmark our proposed method across different architectures and datasets. We further report the qualitative and quantitative results to show the performance of our model.


\vspace{-5 pt}

\subsection{Implementation Details}

We evaluate SaliencyDecor on standard benchmarks, including MNIST~\cite{lecun2002gradient}, CIFAR-10~\cite{krizhevsky2009learning}, and Caltech-101~\cite{fei2004learning}, Birds dataset~\cite{etmann2019connection} and Tiny Imagenet~\cite{deng2009imagenet}. For architectures, we used a 2-layer CNN with 392 feature maps for MNIST, a 2-layer CNN with 512 feature maps for CIFAR-10, ResNet-18 with 512 feature maps for Caltech-101, and ResNet-50 for the Birds dataset. The models were implemented in PyTorch with SGT (momentum = 0.9), an initial learning rate of 0.01 with cosine annealing, and batch size of 128. Hyperparameters were consistent across experiments: decorrelation strength $\lambda=0.01$, consistency weight $\alpha=0.1$, masking ratio $\rho = 25\%$ and group size $G=64$ for ZCA whitening.

\vspace{-5 pt}
\subsection{Results}
\subsubsection{Visual Attribution Analysis}
\begin{figure*}[t]
  \centering
  \includegraphics[width=2\columnwidth]{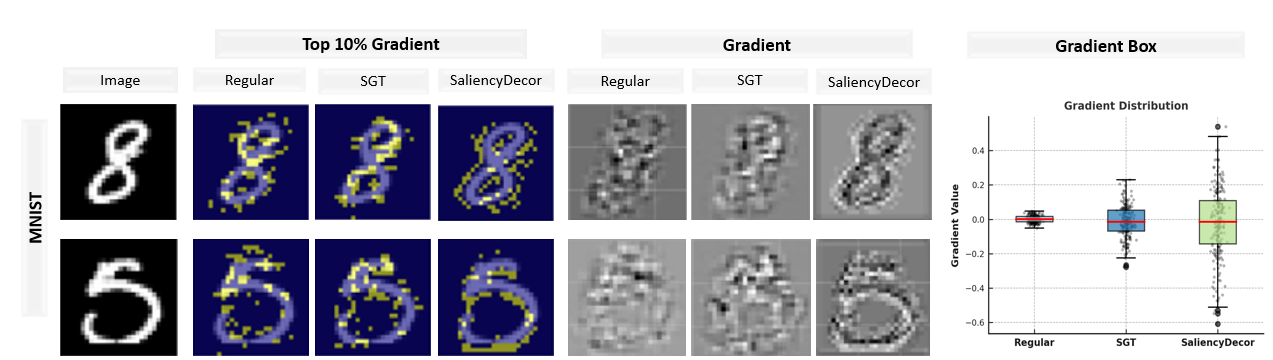}
  \caption{Gradient visualization and distribution analysis for MNIST. Left: Original digit images. Middle columns: Top 10\% gradients and full gradient maps for Regular, SGT, and SaliencyDecor methods. Right: Statistical distribution of gradient values showing box plots. SaliencyDecor produces focused attributions with significantly wider separation between important and non-important features.}
  \label{fig:gradients_mnist}
\end{figure*}

\begin{figure}[h]
  \centering
  \includegraphics[width=0.95\columnwidth]{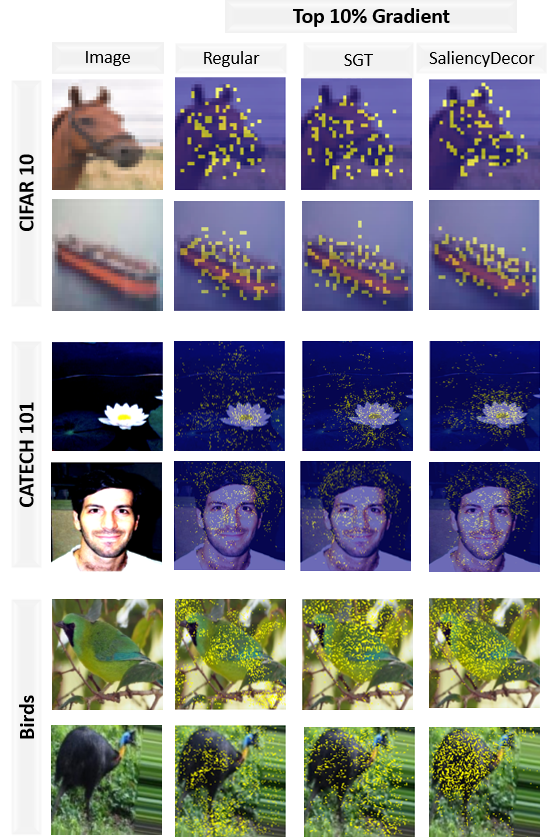}
  \caption{Gradient visualization analysis for complex datasets. Shows original images from CIFAR-10, Caltech-101, and Birds datasets with corresponding gradient patterns from Regular, SGT, and SaliencyDecor methods. SaliencyDecor produces coherent attributions that align with discriminative features while suppressing noise across all complex visual recognition tasks.}
  \label{fig:gradients_complex}
\end{figure}

Figure~\ref{fig:gradients_mnist} compares gradient maps on MNIST for standard training (Regular), SGT~\cite{ismail2021improving}, and SaliencyDecor. Our method produces cleaner, more concentrated attributions aligned with digit structure, while baselines exhibit scattered and noisy gradients. Similar trends are observed on CIFAR-10, Caltech-101, and Birds (Fig.~\ref{fig:gradients_complex}), where SaliencyDecor yields sharper, object-focused saliency and suppresses background noise.

\subsubsection{Statistical Gradient Distribution}

The box plots in Figure~\ref{fig:gradients_mnist} (right) show clear differences in gradient behavior across methods. Standard training yields a collapsed distribution with little separation between important and non-important features, whereas SaliencyDecor produces a wider spread and stronger separation of attribution values, particularly on MNIST.

\subsubsection{Robustness and Faithful Attribution Evaluation}
\begin{figure}[t]
  \centering
  \includegraphics[width=0.89\columnwidth]{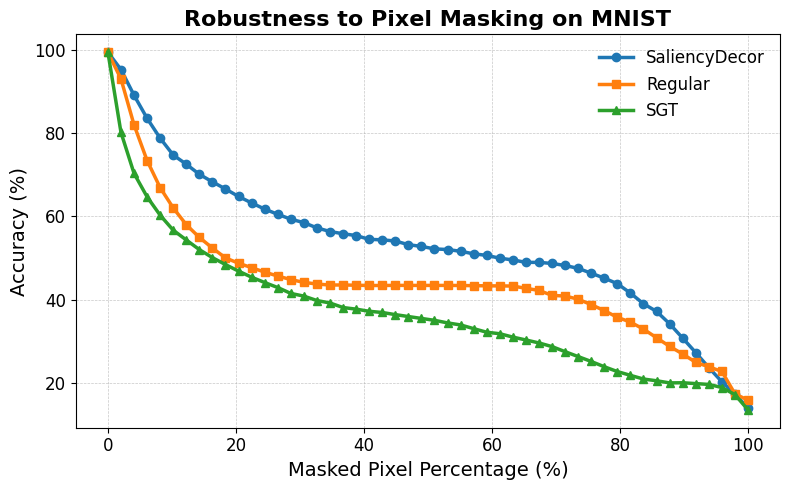}
  \caption{Accuracy degradation under progressive feature masking on MNIST: the baseline model (AUC = 5353.33), the conventional saliency method (AUC = 4458.52), and SaliencyDecor (AUC = 3707.19) show that models trained with SaliencyDecor exhibit significantly steeper accuracy decline, confirming more precise identification of truly important features compared to conventional methods.}
  \label{fig:accuracy_drop}
\end{figure}

We evaluate attribution fidelity using a perturbation-based masking test, where input features are progressively removed according to their saliency scores and the resulting accuracy degradation is measured. Figure~\ref{fig:accuracy_drop} shows the results on MNIST. Models trained with SaliencyDecor exhibit a substantially steeper accuracy decline than both standard training and SGT, indicating that the identified features are more critical for prediction. This effect is most evident between 12\% and 36\% masking, where accuracy drops from approximately 75--80\% to around 60\%, while competing methods maintain higher accuracy over the same range. 
Table~\ref{tab:mnist_results} further quantifies attribution quality using the Area Under the Masking Curve (AUC), where lower values indicate better fidelity. SaliencyDecor achieves the lowest AUC while maintaining competitive classification accuracy, demonstrating improved identification of discriminative features.

\begin{table}[h]
  \centering
  \setlength{\tabcolsep}{6pt}
  \caption{MNIST Performance Comparison (AUC)}
  \label{tab:mnist_results}
  \begin{tabular}{cccc}
    \toprule
    \textbf{Baseline (w/o Saliency)} & \textbf{SGT} & \textbf{SaliencyDecor} \\
    \midrule
    5353.33 & 4458.52 & \textbf{3707.19} \\
    \bottomrule
  \end{tabular}
\end{table}

\begin{table*}[t]
\centering
\small
\caption{Accuracy comparison across common benchmarks for saliency-/gradient-based works. 
``--'' indicates that the corresponding literature does not report results for that dataset.
}
\label{tab:acc_saliency_grad_works}
\begin{tabular}{lcccccc}
\hline
\textbf{Dataset} &
\textbf{MNIST} &
\textbf{CIFAR-10} &
\textbf{CIFAR-10} &
\textbf{Caltech-101} &
\textbf{Birds} &
\textbf{Tiny ImageNet} \\
\textbf{Model} &
Simple CNN &
CNN &
ResNet-18 &
ResNet-18 &
CNN / ResNet &
DeiT-Tiny \\
\hline
Baseline (Standard training without Saliency) & \textbf{99.4} & 73.5 & 94.6 & 94.5 & 93.6 & 72.4 \\
Gradient Regularization (SpectReg) \cite{varga2017gradient} & 97.7 & -- & -- & -- & -- & 50.8 \\
Integrated Gradient Correlation \cite{lelievre2024integrated} & 99.3 & -- & -- & -- & -- & -- \\
SCAAT \cite{xu2023scaat} & -- & -- & 90.5 & -- & -- & -- \\
SGT \cite{ismail2021improving} & 99.3 & 73 & 92.9 & 94.7 & 94.3 & 72.9 \\
SaliencyDecor (Ours) & \textbf{99.4} & \textbf{76.4} & \textbf{95.1} & \textbf{96.2} & \textbf{95.2} & \textbf{73.1} \\
\hline
\end{tabular}

\end{table*}

\subsubsection{Extending SaliencyDecor to Vision Transformers}
We extend SaliencyDecor to Vision Transformers by applying feature decorrelation directly to the CLS token representation produced by the transformer encoder. After the final transformer block, the CLS embedding is extracted and partitioned into fixed-size groups, on which a group-wise ZCA whitening operation is applied to reduce feature correlation while preserving feature orientation.
The decorrelated CLS representation is used solely to compute an auxiliary feature decorrelation loss, whereas classification logits are obtained from the original (non-whitened) CLS embedding. This design decouples predictive performance from the whitening operation, allowing decorrelation to act as a regularizer without altering the classifier decision space. The overall training objective combines standard cross-entropy loss with the decorrelation loss, encouraging the model to learn less redundant yet discriminative representations.
This formulation enables a lightweight integration of SaliencyDecor into transformer architectures without modifying the attention mechanism or introducing inference-time overhead.

\subsection{Ablation Study}
\label{sec:ablation}

\subsubsection{Impact of feature decorrelation without saliency guidance}

Table~\ref{tab:ablation_acc_compact} compares SaliencyDecor against a decorrelation-only baseline using Decorrelated Batch Normalization (DBN)~\cite{huang2018decorrelated}. While DBN significantly degrades performance relative to the standard baseline, SaliencyDecor consistently improves accuracy across both seeds. These results indicate that whitening alone is insufficient, and that the benefits arise from the joint integration of decorrelation with saliency-guided masking and consistency objectives.

\begin{table}[h]
\centering
\small
\caption{Ablation results on CIFAR-10 using ResNet-18 under identical training settings. Test accuracy (\%) is reported.}
\label{tab:ablation_acc_compact}
\setlength{\tabcolsep}{6pt}
\begin{tabular}{lc}
\toprule
\textbf{Method} & \textbf{Test Acc. (\%)} \\
\midrule
Baseline (Regular) & 94.59 \\
SGT & 92.77 \\
DBN & 81.40 \\
\textbf{Ours} & \textbf{95.12} \\
\bottomrule
\end{tabular}
\end{table}

\subsubsection{Effect of masking ratio $k$.}
We study the impact of the masking ratio $k$ used to remove low-saliency features during training. As shown in Table~\ref{tab:ablation_combined} (a), masking $25\%$ achieves the best test accuracy (95.12\%). Increasing the masking ratio to $50\%$ degrades accuracy (94.75\%), suggesting that overly aggressive masking removes informative regions and makes the consistency objective harder to satisfy. Masking $75\%$ partially recovers performance (94.96\%), but remains below $25\%$. Moderate masking provides the best trade-off between enforcing robustness and preserving discriminative evidence.

\subsection{Effect of decorrelation weight $\lambda$}

We vary the decorrelation weight $\lambda$ (Table~\ref{tab:ablation_combined}(b)). A small value ($\lambda=0.001$) yields the best accuracy (95.17\%), while larger values slightly degrade performance, indicating that decorrelation is most effective as a mild regularizer.

\section*{Acknowledgments}
This research was supported by the National Science Foundation under Award \#2233893.







\begin{table}[t]
\centering
\caption{Ablation study on masking ratio $k$ and decorrelation weight $\lambda$.}
\label{tab:ablation_combined}

\begin{minipage}{0.48\columnwidth}
\centering
\small
\begin{tabular}{cc}
\toprule
Masking Ratio ($k$) & Acc. (\%) \\
\midrule
25\% & \textbf{95.12} \\
50\% & 94.75 \\
75\% & 94.96 \\
\bottomrule
\end{tabular}

\vspace{2pt}
\textit{(a) Masking ratio}
\end{minipage}
\hfill
\begin{minipage}{0.48\columnwidth}
\centering
\small
\begin{tabular}{cc}
\toprule
$\lambda$ & Acc. (\%) \\
\midrule
0.001 & \textbf{95.17} \\
0.01  & 95.12 \\
0.1   & 95.02 \\
\bottomrule
\end{tabular}

\vspace{2pt}
\textit{(b) Regularization weight}
\end{minipage}
\end{table}



\section{Conclusion}

This work links representation geometry to interpretability, identifying feature correlation as a key cause of degraded saliency. We introduce SaliencyDecor, which integrates ZCA whitening with saliency-guided training to learn decorrelated features and produce cleaner, more reliable attributions without inference overhead. Across benchmarks, our method consistently improves both attribution fidelity and classification performance, demonstrating that feature decorrelation provides a simple and principled approach to a reliable interpretability.

\bibliographystyle{ieeetr}

\bibliography{main}

\end{document}